\documentclass[10pt,twocolumn,letterpaper]{article}

\usepackage{wacv}
\usepackage{times}
\usepackage{epsfig}
\usepackage{graphicx}
\usepackage{amsmath}
\usepackage{amssymb}
\usepackage{outlines}
\usepackage[accsupp]{axessibility}

\def\wacvPaperID{1967} 

\wacvfinalcopy 

\ifwacvfinal
\usepackage[breaklinks=true,bookmarks=false]{hyperref}
\else
\usepackage[pagebackref=true,breaklinks=true,colorlinks,bookmarks=false]{hyperref}
\fi

\begin{document}

\title{THOR-Net: End-to-end Graformer-based Realistic Two Hands and Object Reconstruction with Self-supervision}

\author{
Ahmed Tawfik Aboukhadra$^{1,2}$ $\;\;\;$  
Jameel Malik$^{1,3}$  $\;\;\;$
Ahmed Elhayek$^{4}$ $\;\;\;$\\
Nadia Robertini$^{1}$ $\;\;\;$
Didier Stricker$^{1,2}$\vspace{7pt}\\
$^{1}$DFKI-AV Kaiserslautern$\;\;$
$^{2}$TU Kaiserslautern$\;\;$
$^{3}$NUST-SEECS Pakistan$\;\;$
$^{4}$UPM Saudi Arabia$\;\;$  \\
}


\maketitle
\thispagestyle{empty}

\begin{abstract}
    
    
    Realistic reconstruction of two hands interacting with objects is a new and challenging problem that is essential for building personalized Virtual and Augmented Reality environments. 
    Graph Convolutional networks (GCNs) allow for the preservation of the topologies of hands poses and shapes by modeling them as a graph. 
    In this work, we propose the THOR-Net which combines the power of GCNs, Transformer, and self-supervision to realistically reconstruct two hands and an object from a single RGB image. 
    Our network comprises two stages; namely the features extraction stage and the reconstruction stage. In the features extraction stage, a Keypoint RCNN is used to extract 2D poses, features maps, heatmaps, and bounding boxes from a monocular RGB image. 
    Thereafter, this 2D information is modeled as two graphs and passed to the two branches of the reconstruction stage. 
    The shape reconstruction branch estimates meshes of two hands and an object using our novel coarse-to-fine GraFormer shape network. 
    The 3D poses of the hands and objects are reconstructed by the other branch using a GraFormer network.
    Finally, a self-supervised photometric loss is used to directly regress the realistic textured of each vertex in the hands' meshes. 
    Our approach achieves State-of-the-art results in Hand shape estimation on the HO-3D dataset (10.0mm) exceeding ArtiBoost (10.8mm). 
    It also surpasses other methods in hand pose estimation on the challenging two hands and object (H2O) dataset by 5mm on the left-hand pose and $1$ mm on the right-hand pose. THOR-Net code will be available at \url{https://github.com/ATAboukhadra/THOR-Net}.

    

\end{abstract}

\begin{figure}[t]
\begin{center}
  \includegraphics[width=\linewidth]{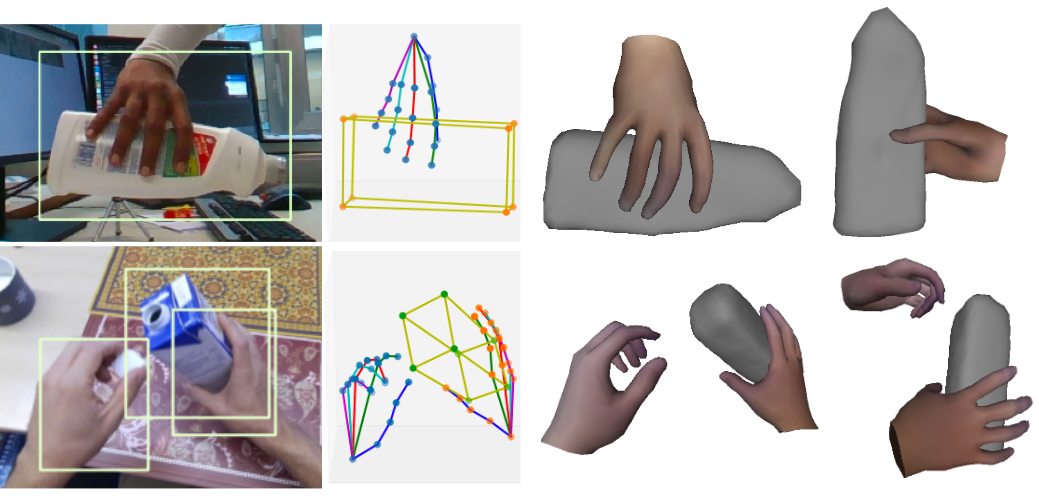}
\end{center}
  \caption{Our Graformer-based algorithm jointly reconstructs up to two hands Poses and textured shapes together with a shape of one object from a monocular RGB image. Note that the hands' textures of the above shapes were directly regressed for each vertex based on self-supervision training. \vspace{-3mm}}
\label{fig:teaser}
\end{figure}

\section{Introduction}

Realistic hands-object shape reconstruction is crucial for many Augmented Reality (AR) and Virtual Reality (VR) applications in order to create a more immersive, personalized experience for the users.
Moreover, the hand pose is useful for human-computer interaction, action recognition, human behavior analysis, and gesture recognition applications \cite{Doosti_2020_CVPR, HandVoxNet++2021, Almadani2021, chao:cvpr2021, rashed, airsig}.
The recent advancements in hand, body and object pose estimation \cite{hasson19_obman, Park_2022_CVPR_HandOccNet, Hampali_2022_CVPR_Kypt_Trans, hasson2020leveraging, HandVoxNet++2021} are promising.
However, few attention is given to the joint reconstruction of two hands interacting with an object \cite{hasson2020leveraging, hasson19_obman, Almadani2021, Doosti_2020_CVPR, tekin2019h+}.
This is a challenging problem due to varying hand shapes, texture, many degrees of freedom (DOF), self-similarity of hands parts, two-hands self-occlusions, and hand-object mutual occlusion, especially from a monocular RGB image as it only contains 2D information.

By utilizing the recent advances in deep learning (e.g., GCNs, Transformers, and self-supervised learning), several algorithms for simultaneous hand pose and shape estimation have been introduced. 
Recently, many researchers used Graph Convolutional Networks (GCNs) \cite{kipf2016semi} to address the challenges of pose estimation  \cite{Doosti_2020_CVPR, yan2018spatial, ZhaoGraFormer, cai2019exploiting} and shape reconstruction \cite{Almadani2021, HandVoxNet++2021, wang2018pixel2mesh}. 
GCNs preserve the inherent kinematic and graphical structure of hand pose and shape. 
This feature allows GCNs to handle depth ambiguity and occlusions as it correlates the visible parts of the hand with the non-visible parts \cite{Doosti_2020_CVPR}.
Transformer networks \cite{VaswaniTransformer} have also shown great abilities in many domains such as NLP \cite{devlin2018bert}.
Transformers have shown to be highly effective in many Computer Vision domains \cite{dosovitskiy2020vit}. 
Many researchers have studied the effectiveness of Transformers in hand pose and shape estimation \cite{huang2020hot, Park_2022_CVPR_HandOccNet, ZhaoGraFormer, Hampali_2022_CVPR_Kypt_Trans, xu2022vitpose, lin2021end-to-end}.


In this paper,  we propose the first —to the best of
our knowledge— approach with GCNs, Transformers, and self-supervision which simultaneously estimates the 3D shape and the 3D pose of two hands interacting with an object together with the texture of each vertex of the hands given a monocular RGB image as shown in Figure \ref{fig:teaser}.

THOR-Net is based on Keypoint RCNN which extracts several 2D features (i.e., heatmaps, bounding boxes, features maps, and 2D pose) from the monocular RGB image. 
To benefit from the power of the GCNs we model all this 2D information as two graphs.
One graph is passed through our novel coarse-to-fine GraFormer shape generator network to estimate meshes for the hands and the object. 
This network gradually increases the number of nodes in the graph starting from the pose until reaching the shape while gradually decreasing the size of the features to only 3 values (x,y,z) that correspond to each vertex location in 3D space.
The other graph is passed through a 2D-to-3D pose estimation network which is based on GraFormer to estimate 3D poses for the hands and object. 

The hands' textures of the meshes are directly regressed by using a self-supervision photometric loss. 
To this end, the texture of each vertex is learned by orthographic projection to the input image. 
In contrast to HTML \cite{HTML_eccv2020} which learns the statistical hand texture model from a limited set of hand texture samples, our photometric loss approach allows for learning hand textures from a huge set of RGB images of any hands dataset.


To summarize, we make the following contributions:

\begin{outline}
    \1 A novel pipeline to reconstruct a realistic 3D shape for two hands and objects from RGB images with the following novelties:
        \2 Utilizing heatmaps and features produced by the Keypoint RCNN to build graphs that help our GraFormer-based networks to estimate 3D pose and shape. 
        \2 Proposing a coarse-to-fine GraFormer for two hands and object reconstruction. 
        
    \1 Applying self-supervision based on photometric loss to give a more realistic view of hands. 
    \1 Our method achieves state-of-the-art results for hand mesh estimation on HO-3D (v3) and hand pose estimation on the H2O dataset as shown in Section \ref{sec:experiments}.
\end{outline}


\section{Related Work}

Although most of the existing works focus on the reconstruction of a single interacting hand, our work addresses a more challenging problem of two hands and object reconstruction. Here, we briefly describe the most related works.

\subsection{GCNs for Pose Estimation}

%
%

Recently, 3D pose estimation from 2D pose using Graph Convolutional Networks (GCNs) showed very promising results \cite{Doosti_2020_CVPR, ZhaoGraFormer}. 
Using a single Keypoint from the 2D pose to estimate its counterpart in 3D is a nondeterministic problem. 
However, using the information about other 2D keypoints and their relation to the target keypoint can be useful to estimate its 3D location.
%
%
The authors of the HopeNet \cite{Doosti_2020_CVPR} introduced an adaptive GraphUNet that pools the 2D pose in five stages, and then unpools it to get the 3D pose while having skip connections between the corresponding pooling and unpooling layers. 

The GraFormer \cite{ZhaoGraFormer} transforms 2D poses to 3D, however, it shows a much better performance than the HopeNet because of combining Graph Convolutional layers with the Transformer \cite{VaswaniTransformer} and attention mechanism. The GraFormer is able to extract local features from the nodes using graph convolutional layers and also extract global information about the entire graph using the attention layers. 

The spatiotemporal graph solves the depth ambiguity and severe occlusion challenges in 3D pose estimation \cite{cai2019exploiting, yan2018spatial}. 
Temporal continuity in videos imposes temporal constraints \cite{hasson2020leveraging}.
Therefore, Cai \textit{et al.} \cite{cai2019exploiting} created a Spatio-temporal graph from a few temporally adjacent 2D body poses by creating additional edges between the joints and their counterparts in neighboring frames. 



\subsection{Hand-Object Reconstruction}
Most of the existing works focus on hand shape estimation under interaction with an object without considering object shape reconstruction. 
The Keypoint Transformer \cite{Hampali_2022_CVPR_Kypt_Trans} achieves state-of-the-art results in hand pose estimation from RGB images by extracting features from the image for each keypoint and correlating those features using self-attention layers. 
HandOccNet \cite{Park_2022_CVPR_HandOccNet} is a very recent and robust transformer-based model that solves the ambiguity of occlusions between hands and objects by injecting features from visible areas of the hand to areas where the hand is occluded by the object. 
ArtiBoost \cite{li2021artiboost} aims to solve the lack of diversity of hand-object poses within the 3D space in any hand-object dataset by creating synthetic images. They use both synthetic and real images to train a CNN regression model that estimates the pose. 
Liu \textit{et al.} \cite{liu2021semi} leveraged spatiotemporal consistency in RGB videos to generate labels for semi-supervised training to estimate 3D pose.
%
%

\begin{figure*}[t]
\begin{center}
  \includegraphics[width=\linewidth]{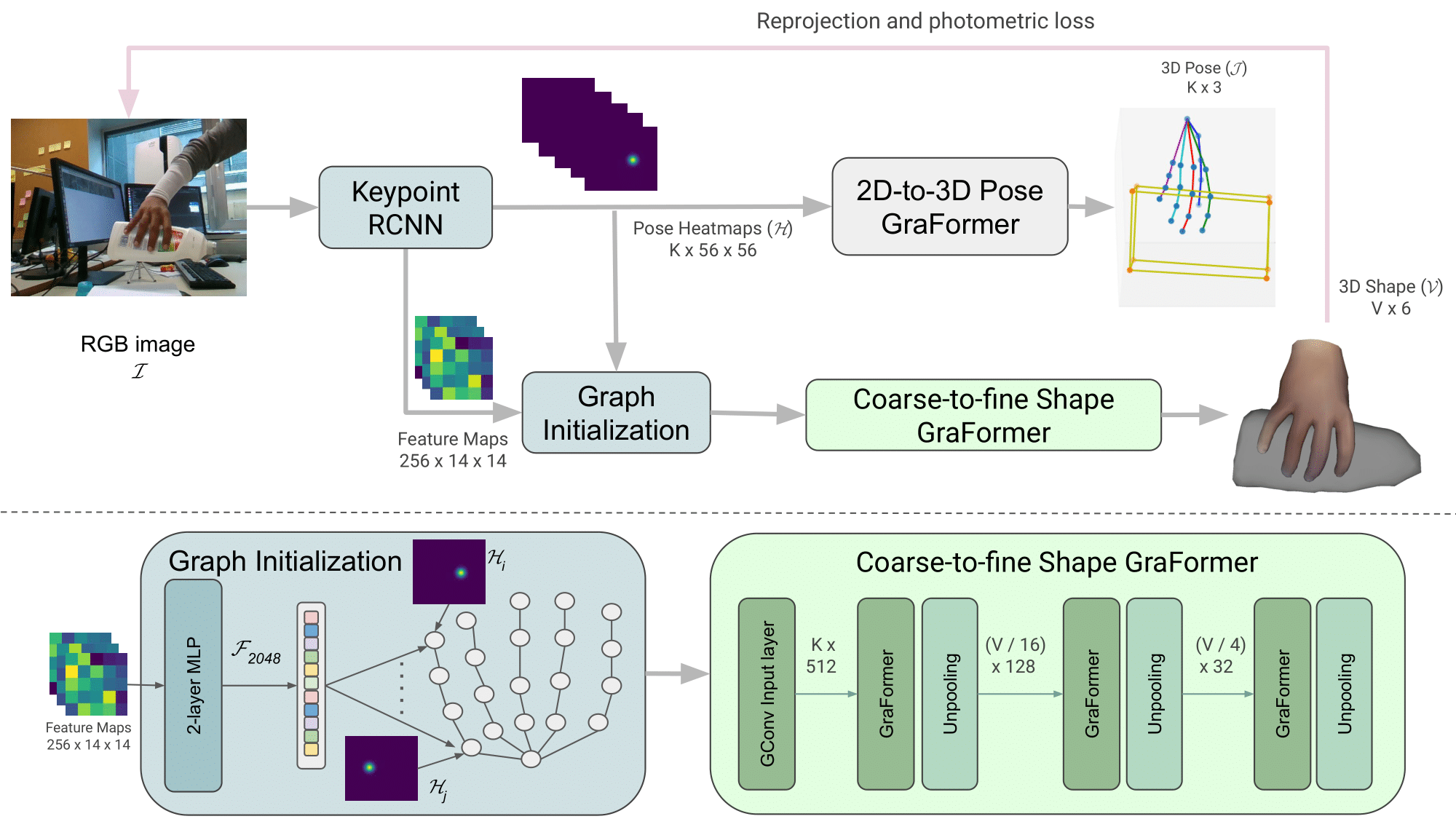}
\end{center}
  \caption{An overview of our approach to estimating the 3D pose and 3D shape for hands interacting with an object from a monocular RGB frame. \textrm{K} is the number of keypoints in the Pose and \textrm{V} is the number of vertices in the Shape. The lower part describes more details about the graph initialization and the Coarse-to-fine Shape GraFormer network. \vspace{-3mm}}
\label{fig:pipeline}
\label{fig:onecol}
\end{figure*}

Two-hands and object reconstruction does not get enough attention compared to hand-object pose estimation and hands-only reconstruction. 
Hasson \textit{et al.} \cite{hasson2020leveraging} used a network that outputs the MANO \cite{mano} parameters for the hand and the object class with its 3D transformation parameters. 
One important aspect of their work is that they use photometric consistency over time as a semi-supervised training scheme when some frames are not annotated.
%
%
In their follow-up work \cite{hasson20_homan}, Hasson \textit{et al.} first detect and segment hands and objects within an RGB image. After that, they estimate hand shape and object pose and optimize them using  loss terms for smoothness and collision. 

Malik \textit{et al.} \cite{s19173784, HandVoxNet2020} investigated hand pose and shape estimation from depth maps.
\cite{HandVoxNet2020,HandVoxNet++2021} used voxelized depth maps to estimate a voxelized shape and a shape surface for the hand, followed by a registration step. 
EventHands \cite{rudnev2021eventhands} is a network that uses an Event camera input to capture and reconstruct hand motions of unprecedented speed.
Almadani \textit{et al.} \cite{Almadani2021} created a depth-based coarse-to-fine hand object reconstruction network that is built on the GCN HopeNet \cite{Doosti_2020_CVPR}. 
After evaluating different input modalities for their model, they show that a voxelized representation of a depth map and a corresponding RGB image is the best input modality.
Pixel2Mesh \cite{wang2018pixel2mesh} is a GCN network that estimates 3D shapes for objects from monocular RGB frames. 

%
%

\section{Method} 

The proposed pipeline is shown in Figure \ref{fig:pipeline}, it uses RGB frame $\mathcal{I}$ as an input and predicts the target 3D pose $\mathcal{J}$ and the 3D shape $\mathcal{V}$ for hands and objects. 

%

%
%
\subsection{Keypoint RCNN}

Mask RCNN \cite{HeMaskRCNN} is effective object detection and semantic segmentation model built over Faster RCNN \cite{RenFasterRCNN}. 
Mask RCNN proposes Regions-of-interest (RoIs) within an image that contains objects and estimates the bounding box and the class for those objects. 
The authors of the Mask RCNN created a variant called Keypoint RCNN that estimates heatmaps of the location of any set of 2D keypoints within the RoI. 
For every keypoint there is a heatmap of the location of that keypoint. 
From bounding boxes and heatmaps, Keypoint RCNN can estimate 2D locations in the image that compose a 2D pose. 
We train the Keypoint RCNN to estimate the 2D pose for hands and objects knowing the projection of the 3D pose to 2D. 
Hence, the Keypoint RCNN provides important information from RGB images such as bounding boxes for hands and objects, heatmaps for keypoints within those boxes, and RoI features. 

To train the Keypoint RCNN, bounding box annotations are required. 
To obtain the bounding boxes, we use the 2D projection of the 3D pose to the image. 
The minimum x, and y values, and the maximum x, and y values of the 2D pose are considered to be the bounding box. 
Figure \ref{fig:pipeline} shows the two outputs of the Keypoint RCNN which are the Heatmaps and the features, that are used to train our models for 3D pose and shape estimation.
One important advantage of the Keypoint RCNN's ability to localize objects in raw images removes the need for image preprocessing such as cropping the hand-object region.

%


\subsubsection{Feature Extractor}

The backbone of the Keypoint RCNN consists of a ResNet50 \cite{resnet} and a Feature Pyramid Network (FPN) \cite{resnet} that produces multi-scale features for the RGB image. 
The backbone produces RoI-specific features after passing the multi-scale features into a multi-scale RoI align layer \cite{HeMaskRCNN}. 
This allows us to capture custom features for the RoI that contains hands or objects. 
We use those features to enrich the nodes of the coarse-to-fine GraFormer.
To compress the features before passing them to the next stage, the RCNN passes the features to a 2-layer MLP that produces a compressed $2048$ feature vector for each RoI. 
These feature vectors are appended to the heatmaps to produce a graph representation for the shape generator, as described in Sec. \ref{sec:ShapeGraformer}.

%
%
%
%
%
%

%

\subsection{Pose GraFormer}
%
%

\begin{figure}[t]
\begin{center}
  \includegraphics[width=\linewidth]{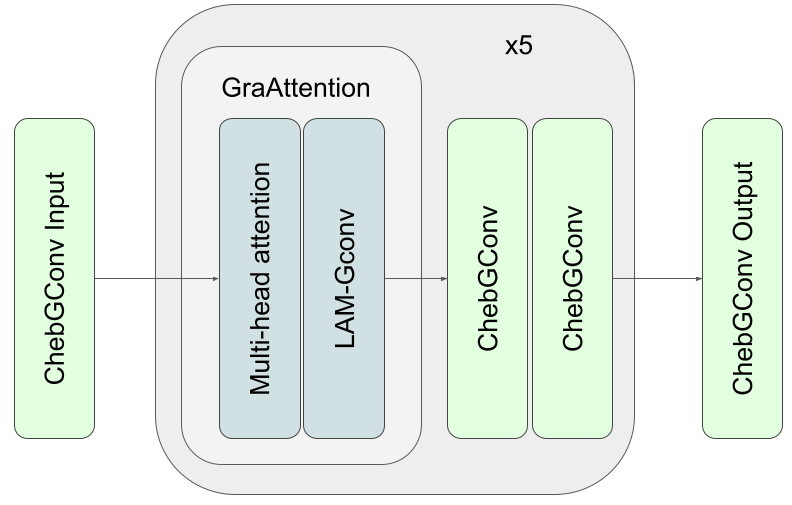}
\end{center}
\vspace{-3mm}
  \caption{An illustration of the GraFormer network. The network consists of GCN and Attention layers repeated multiple times. \vspace{-4mm}}
  \label{fig:graformer}
\end{figure}

To convert the 2D information extracted by the Keypoint RCNN into 3D space, we use the GraFormer. 
The GraFormer \cite{ZhaoGraFormer} is a Graph Neural Network that is designed to utilize the advantages of Graph Convolutional layers and Attention layers \cite{VaswaniTransformer}. 
Graph Convolutional layers extract features from graph-like data depending on the connectivity between the nodes, in our case, the edges between adjacent keypoints of the pose or the shape.
Furthermore, the multi-head self-attention \cite{VaswaniTransformer} layers within the GraFormer extract global features from the graph.
Both concepts caused the GraFormer to outperform other methods in 2D-to-3D pose lifting. 

As illustrated in Figure \ref{fig:graformer}, the GraFormer consists of a GraAttention layer that is a multi-head self-attention layer with 4 heads. 
The last layer of the GraAttention is a LAM-GConv layer which is a graph convolution layer with a trainable adjacency matrix. 
The GraAttention is followed by $2$ layers of a special type of graph convolutions called ChebGConv\cite{DefferrardChebGConv} composing the main building component of the GraForemr. 
This component of the GraAttention and the ChebGConv is repeated five times to create the GraFormer. 
The first use of the GraFormer in our work is to transform the heatmaps of the Keypoint RCNN to 3D pose coordinates of the hand and the object. 
Instead of representing every node in the graph using its 2D pixel location, we found empirically that using heatmaps is more accurate. 

%
%

%
%
\subsection{Coarse-to-fine Shape GraFormer}
\label{sec:ShapeGraformer}

To generate the 3D shape, we propose a coarse-to-fine GraFormer that gradually increases the number of vertices starting from a 2D pose graph and ending with the 3D shape. 
Almadani \textit{et al.} \cite{Almadani2021} and Wang \textit{et al.} \cite{wang2018pixel2mesh} previously explored Coarse-to-fine GCNs to generate 3D shapes. 
However, given the improved performance of GraFormer compared to normal GCNs, we replace their suggested Graph convolutional layers with GraFormers as shown in Fig. \ref{fig:pipeline}.

The network consists of three stages and each stage is composed of a GraFormer followed by an unpooling layer that increases the number of nodes in the graph. 
The input graph to the Coarse-to-fine GraFormer consists of $29$ nodes. 
Every node \textit{i} holds the feature vector $\mathcal{F}_{2048}$ and the corresponding heatmap $\mathcal{H}_{i}$ of size $56\times56$ as shown in Fig. \ref{fig:pipeline}. 
After flattening the heatmap and appending it with $\mathcal{F}_{2048}$, the size of the node representation is $5184$.

To model a hand mesh as a graph, we use the MANO \cite{mano} faces to create the adjacency matrix. 
However, there are two challenges to creating such a coarse-of-fine graph network. 
The objects in both HO-3D and H2O datasets have a varying number of vertices and they do not have a consistent topology. 
The second challenge is that the intermediate graph layers within the coarse-to-fine network require a simplified version of the adjacency matrix as they have a lower number of vertices in their graphs. 
In Section \ref{sec:simplification}, we describe how to create simplified versions of the hand mesh adjacency matrix. 
In section \ref{sec:objTopology}, we describe how to create a consistent topology for the objects.

%
%
%
%
%



\subsubsection{Hand Mesh Downsampling}
\label{sec:simplification}
%
%
To create an intermediate graph representation of the hand, we use the Quadric Edge Collapse Decimation algorithm (QECD) \cite{garland1997surface, pymeshlab} to downsample the default MANO hand mesh. The faces of the resulting simplified mesh create the adjacency matrix of the intermediate graphs. We simplify the $778$ hand vertices to two granularity levels (i.e., $49$ and $194$) to correspond to levels $1$ and $2$ in the coarse-to-fine network respectively.

\subsubsection{Object Topology}
\label{sec:objTopology}
%
%
To solve the problem of the inconsistent object topology, we use a trainable approach from PyTorch3d \cite{ravi2020pytorch3d} to deform a sphere with a constant topology to every object. 
An Icosphere is a sphere that results from the recursive subdivison of the polygons of a $20$-faces polyhedron \cite{pymeshlab}. 
At level $4$ of subdivision, the Icosphere has $2556$ vertices. 
We use the QECD algorithm to simplify that sphere to $1000$ vertices. 
This step is executed to control the number of vertices that represent the object shape depending on the complexity of the model and the required reconstruction quality.

%
%
%
%
%
%

\begin{figure}[t]
\begin{center}
  \includegraphics[width=\linewidth]{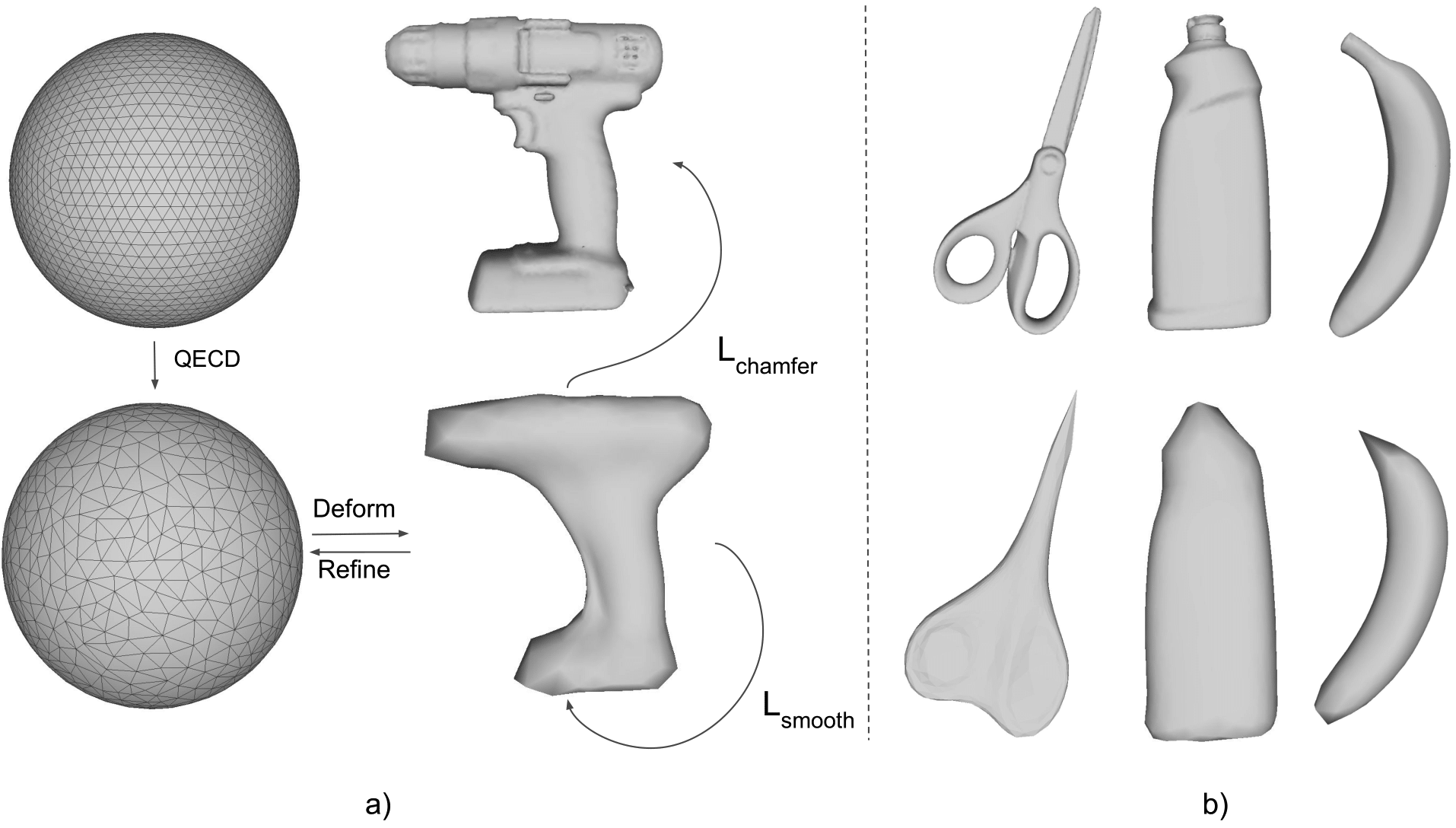}
\end{center}
\vspace{-3mm}
  \caption{a) The process of simplifying and deforming a sphere to obtain a consistent topology representation for the mesh. b) Examples of different 3D object models and their simplified versions. The bottom row shows the downsampled spheres.  \vspace{-4mm}}
\label{fig:objects}
\end{figure}

To learn the displacement of every vertex in the sphere to the target object mesh, the deformation algorithm minimizes the Chamfer distance $\mathcal{L}_{chamfer}$  between the deformed sphere and the target mesh as shown in Figure \ref{fig:objects}. Along with the Chamfer loss, 3 additional regularization losses add smoothing effects on the resulting deformed sphere. The three losses are Edge length $\mathcal{L}_{edge}$, normal consistency of neighboring faces $\mathcal{L}_{norm}$ and Laplacian smoothing $\mathcal{L}_{laplacian}$. The final loss term for deformation is:
\begin{equation}
    \mathcal{L} = \mathcal{L}_{chamfer} + \mathcal{L}_{edge} + \lambda_1 * \mathcal{L}_{norm} + \lambda_2 * \mathcal{L}_{laplacian}
\end{equation}

$\lambda_1$ equals $0.01$ and $\lambda_2$ equals $0.1$.
An SGD optimizer minimizes the weighted sum of the aforementioned losses until the sphere reaches the closest state to the target object. Figure \ref{fig:objects} shows some of the objects in the YCB dataset \cite{ycb}, and their corresponding deformed sphere with $1000$ vertices.

%
%

\subsubsection{Photometric Loss}
\label{sec:photo}
%
%
%
%
Estimating a texture value for every vertex in the mesh gives it a richer representation of a personalized hand shape and a more realistic view \cite{HTML_eccv2020}. 
In addition, it helps to improve the alignment between the estimated shape with the target reprojection improving the reconstruction error.
Furthermore, exploring hand textures is an interesting problem in the field of VR and AR as it improves the immersive experience.
Qian \textit{et al.} \cite{HTML_eccv2020} proposed the first parametric hand texture model (HTML) for the reconstruction of realistic hand texture. 
Although this model allows the generation of a diverse set of hand textures by randomly sampling from the texture parameters, this approach is limited by the small training dataset (i.e., $51$ subjects) which is used to build the hand texture model. 
This means that the proposed statistical model can not represent any texture that is not covered in this dataset. 
In this paper, we propose a direct texture regression approach that is based on self-supervision using photometric loss \cite{hasson2020leveraging, HTML_eccv2020}. 
To this end, a texture is directly learned for each hand mesh vertex together with the 3D position of this vertex. 
In contrast to HTML which learns the statistical hand texture model from a limited set of hand texture samples, our approach allows for learning hand textures from a huge set of RGB images of any hand dataset.
From this motivation, we add an additional loss term to train the model, and instead of estimating only XYZ for the vertices, the model estimates an RGB value as well. 

To calculate this loss, the target 3D shape is first projected into the image using the camera intrinsics. 
After that, the corresponding pixel RGB values of the projected vertices are penalized with the RGB values that the model estimates by calculating the MSE between both.
This justifies the six values, as shown in Figure \ref{fig:pipeline}, for the final shape. 
The photometric loss $\mathcal{L}_{photo}$ is defined as follows:
\begin{equation}
    \mathcal{L}_{photo} = MSE(\mathcal{I}[\textit{proj}(\mathcal{V}_{gt})], \mathcal{V}_{pred, rgb})    
\end{equation}

\section{Experiments}
\label{sec:experiments}

In this section, we discuss the datasets and the implementation details for each dataset. 
After that, we discuss the training details and the loss functions.
We then report and compare our results quantitatively and qualitatively with other methods in sections \ref{sec:pose} and \ref{sec:shape}.
Further, we conduct an ablation study to show the effectiveness of the components of our pipeline. 

\begin{figure*}[t]
\begin{center}
  \includegraphics[width=0.9\linewidth]{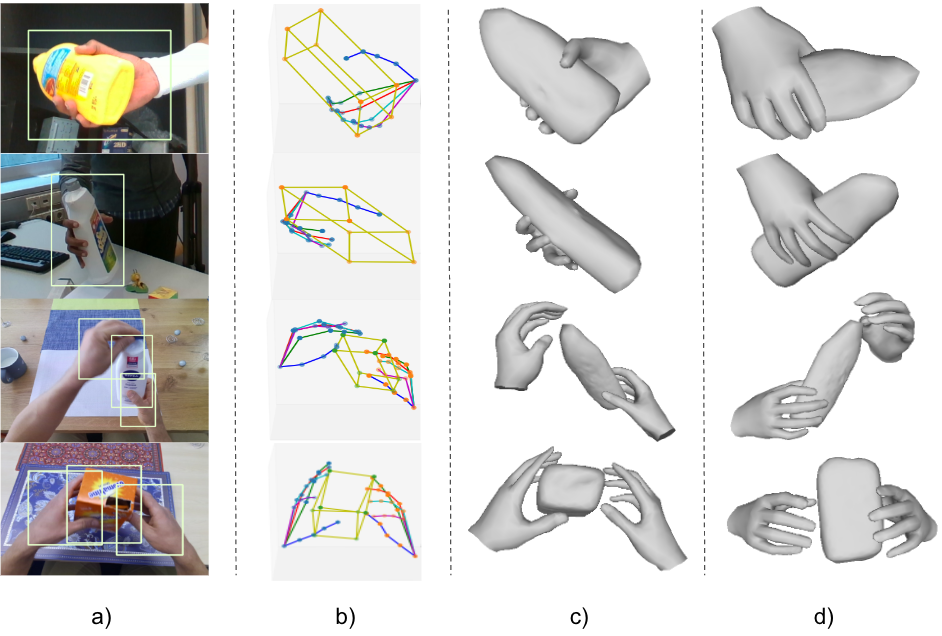}
\end{center}
\vspace{-3mm}
  \caption{Qualitative results for Hand-Object pose and shape estimation from the HO-3D and H2O datasets. a) Input frame with the predicted bounding boxes. b) 3D pose estimation. c) \& d) Two views for the 3D reconstructed hand-object interaction.  \vspace{-4mm}}
\label{fig:qualitative}
\end{figure*}

\subsection{Datasets and Implementation Details}

Researches have created many datasets recently to model the markerless hand-object interactions such as HO-3D \cite{hampali2020honnotate}, H2O \cite{Kwon_2021_ICCV}, H2O-3D \cite{Hampali_2022_CVPR_Kypt_Trans}, DexYCB \cite{chao:cvpr2021}, FPHAB \cite{FirstPersonAction_CVPR2018} and ContactPose \cite{Brahmbhatt_2020_ECCV}. 
We evaluate our method on two recent public benchmark datasets: HO-3D \cite{hampali2020honnotate} and H2O \cite{Kwon_2021_ICCV}. 
HO-3D dataset has one hand interacting with an object while the H2O has two hands interacting with an object.
%
%


%
%


The HO-3D video dataset \cite{hampali2020honnotate} contains 3D pose annotations for hand and a manipulated object under severe occlusions. 
The dataset also contains annotations for the MANO parameters and object labels. 
From the PyTroch Mano model\cite{hasson19_obman}, we acquire the hand vertices and faces and apply simplification to the hand to acquire the intermediate adjacency matrices for the coarse-to-fine network as described in section \ref{sec:simplification}.
All the $10$ objects within the dataset were acquired from the YCB dataset \cite{ycb}. 
We create a spherical representation for the objects following the description in \ref{sec:objTopology}, and use the pose to transform them into the 3D camera space.
All the 3D points are translated such that the palm of the hand is used as the origin of 3D space.
To train the Keypoint RCNN for HO-3D, we consider the hand and the object to be inside the same bounding box as shown in Figure \ref{fig:qualitative}.
We report the results on the second and third versions of the dataset.
The number of keypoints in the 3D pose $\mathcal{J}$ is $29$; $21$ for the hand joints and $8$ for the object corners.
The number of vertices in the 3D shape $\mathcal{V}$ is $1778$; $778$ for the hand mesh and $1000$ for the object mesh.


H2O \cite{Kwon_2021_ICCV} is a new benchmark video dataset that contains the 3D pose annotations for two hands and an object along with the MANO parameters of the hands and the label of the object. 
The dataset covers $8$ objects and provides a 3D model for all of them. 
We follow the same approach used with HO-3D to acquire the 3D mesh of the hands and the object in H2O.
The dataset was captured from five different camera views, however, in this work, we only focus on the egocentric view.

To train the Keypoint RCNN for H2O, we separate each hand's bounding box from the object bounding box. 
We use the 2D keypoints produced by the Keypoint RCNN as an input to the GraFormer instead of the heatmaps as it has shown better results on the H2O dataset.
The number of keypoints in the 3D pose $\mathcal{J}$ is $50$; $21$ for each hand's joints and $8$ for the object corners.
The number of vertices in the 3D shape $\mathcal{V}$ is $2556$; $778$ for each hand's mesh and $1000$ for the object mesh.



\subsection{Training}

To train the THOR-Net, five losses are required:
%
%
Cross Entropy Loss for heatmaps $\mathcal{L}_\mathcal{H}$, Bounding box classification $\mathcal{L}_{cls}$, Mean-squared Error (MSE) for bounding box estimation $\mathcal{L}_{bb}$, MSE to penalize the 3D pose $\mathcal{L}_\mathcal{J}$ and MSE to penalize the 3D shape $\mathcal{L}_\mathcal{V}$. We train our network with a combined loss function:

\begin{equation}
    \mathcal{L} = \mathcal{L}_\mathcal{H} + \mathcal{L}_{cls} + \mathcal{L}_{bb} + \mathcal{L}_\mathcal{J} + \mathcal{L}_\mathcal{V}
\label{eq:loss}
\end{equation}

In the case of generating textured shapes, we add the $\mathcal{L}_{photo}$ as described in Section \ref{sec:photo}.
The network has 192M parameters and we train the model using an Adam optimizer \cite{kingma2014adam}, $0.0001$ learning rate, and batch size $8$ on an NVIDIA's A100 GPU.

\begin{figure}[t]
\begin{center}
  \includegraphics[width=\linewidth]{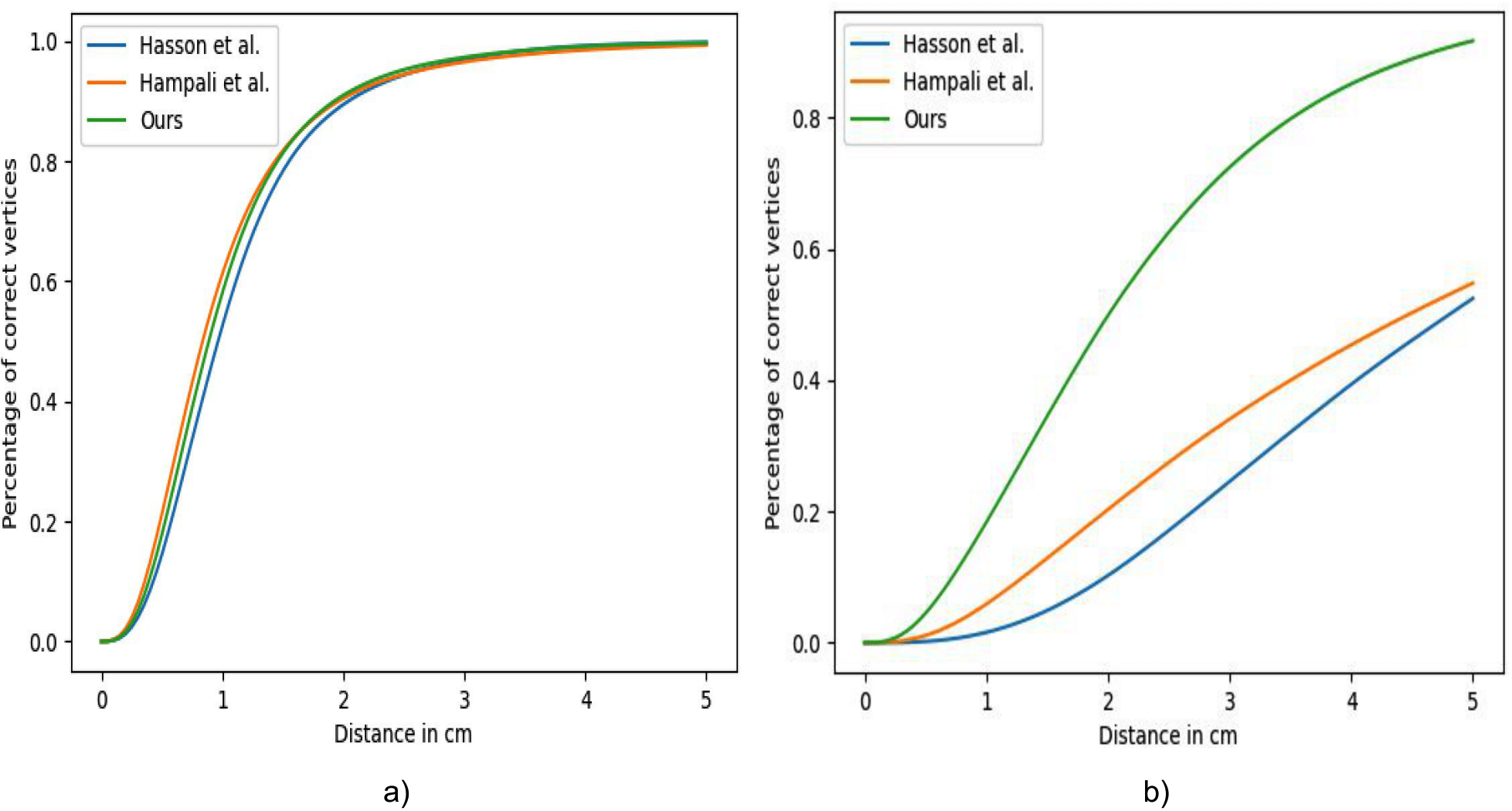}
\end{center}
\vspace{-3mm}
  \caption{PCV over distance in comparison with other methods. a) Procrustes-aligned error. b) Non-aligned error. \vspace{-4mm}}
\label{fig:plots}
\end{figure}

\subsection{Evaluation metrics}

We report the Procrustes aligned and non-aligned MPJPE (Mean Per Joint Position Error) in \textit{mm} on the HO-3D hand pose and shape and compare them with other methods in Tables \ref{tab:ho3d_pose} \& \ref{tab:ho3d_shape}. 
Furthermore, we show the percentage of correct vertices (PCV) over distance in Figure \ref{fig:plots}.
To evaluate our model on the H2O dataset, we report the non-aligned MPJPE of the pose and shape for both hands and object in Table \ref{tab:h2o_pose}.
Finally, we qualitatively show our results in Figure \ref{fig:qualitative}.

\subsection{Evaluation of 3D Pose Estimation}
\label{sec:pose}

We evaluate our method on the two versions of the HO-3D dataset and report the error in \textit{mm} of the hand pose in Table \ref{tab:ho3d_pose}. 
The table contains a comparison of the Procrustes aligned and non-aligned errors with the existing methods.
The reported results can be found on the HO-3D challenge website \footnote{\url{https://codalab.lisn.upsaclay.fr/competitions/4393}}.
%
%
We also evaluate our method for pose estimation on the H2O egocentric view and report the mean joint error for both hands and the object 3D pose in Table \ref{tab:h2o_pose}. 
The table shows the improvement in left and right-hand pose errors compared to previous methods.
The results show an improvement of 5\textit{mm} in left-hand pose estimation and 1\textit{mm} in right-hand pose estimation compared to previous methods.
%
%
The reported results can be found on the H2O challenge website \footnote{\url{https://codalab.lisn.upsaclay.fr/competitions/4822}}.
Qualitative results of the 3D pose on samples from both datasets are shown in Figure \ref{fig:qualitative}.
From the quantitative and qualitative evaluation, our pose estimation method requires future improvements as it does not exceed previous methods on HO-3D, and object pose estimation is not accurate on H2O.
%
%


\begin{table}[t]
\begin{center}
\begin{tabular}{|l|c|c|}
\hline
Methods & $\mathcal{J}$ Al. Err. & $\mathcal{J}$ Err.  \\
\hline\hline
Hasson \textit{et al.} \cite{hasson2020leveraging} & 11.4 & 55.2 \\
Hasson \textit{et al.} \cite{hasson19_obman} & 11.1 & - \\
Hampali \textit{et al.} \cite{hampali2020honnotate} & 10.7 & 84.2 \\
METRO \cite{lin2021end-to-end} & 10.4 & - \\
Liu \textit{et al.} \cite{liu2021semi} & 10.2 & - \\
HandOccNet \cite{Park_2022_CVPR_HandOccNet} & \textbf{9.1} & - \\
THOR-Net (Ours) & 11.3 & 26.3 \\
\hline
Keypoint Trans. \cite{Hampali_2022_CVPR_Kypt_Trans} & 10.9 & - \\
ArtiBoost \cite{li2021artiboost} & \textbf{10.8} & 22.6 \\
THOR-Net (Ours) & 11.2 & 25.6 \\
\hline
\end{tabular}
\end{center}
\caption{\textbf{Comparison with state-of-the-art methods for 3D hand pose estimation on the HO-3D (v2) (upper table) and (v3) (lower table). Shown results are the Procrustes-aligned and non-aligned errors in \textit{mm}.}}
\label{tab:ho3d_pose}
\end{table}


\begin{table}[t]
\begin{center}
\begin{tabular}{|l|c|c|c|}
\hline
Methods & L $\mathcal{J}$ Err. & R $\mathcal{J}$ Err. & Obj. $\mathcal{J}$ Err. \\
\hline\hline
Hasson \textit{et al.} \cite{hasson2020leveraging} & 39.6 & 41.9 & 66.1 \\
H+O \cite{tekin2019h+} & 41.4 & 38.9 & 48.1 \\
H2O \cite{Kwon_2021_ICCV} & 41.5 & 37.2 & \textbf{47.9} \\
THOR-Net (Ours) &  \textbf{36.8} & \textbf{36.5} & 73.9 \\
\hline
\end{tabular}
\end{center}
\caption{\textbf{Comparison with the state-of-the-art methods for 3D pose estimation on H2O dataset. The shown results from left to right are the non-aligned errors in \textit{mm} for the left-hand pose, the right-hand pose, and the object pose.}}
\label{tab:h2o_pose}
\end{table}

\subsection{Evaluation of 3D Shape Estimation}
\label{sec:shape}

We evaluate our method for 3D shape estimation on two datasets and report the results in Table \ref{tab:ho3d_shape}. 
The results show that our Procrustes-aligned mesh error is 10mm on the HO-3D (v3) while the best-found method achieves 10.8mm. 
To the best of our knowledge, we are the first to provide shape evaluations on the H2O dataset.
The left-hand shape error is 54.1mm, the right-hand shape error is 59.4mm and the object shape error is 66.6mm.

\begin{figure}[t]
\begin{center}
  \includegraphics[width=\linewidth]{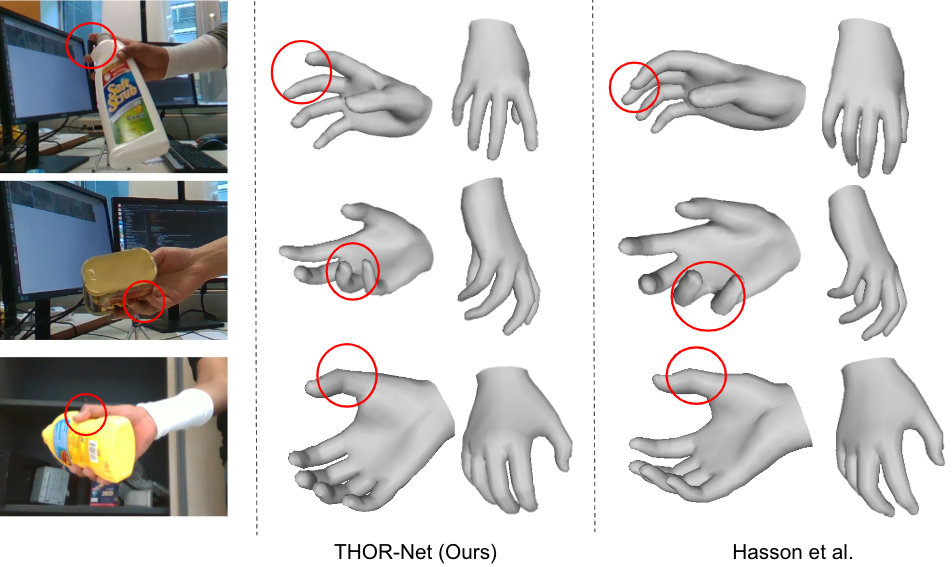}
\end{center}
\vspace{-3mm}
  \caption{Qualitiative comparison with Hasson \textit{et al.} \cite{hasson2020leveraging}. Our method captures better hand details.}
  \label{fig:comparison}
\end{figure}

Qualitative results of the hands-object shapes can be found in Figure \ref{fig:qualitative}.
We also compare our hand shape results with Hasson \textit{et al.} \cite{hasson2020leveraging} in Figure \ref{fig:comparison}.
The results show that our model captures fine hand details. However, the model lacks the ability to generalize to unseen objects.

%
%

%

%
%


\begin{table}[t]
\begin{center}
\begin{tabular}{|l|c|c|}
\hline
Methods & $\mathcal{V}$ Al. Err. & $\mathcal{V}$ Err.  \\
\hline\hline
HandVoxNet++ \cite{HandVoxNet++2021} & - & 27.0 \\
Hasson \textit{et al.} \cite{hasson2020leveraging} & 11.4 & 55.2 \\
METRO \cite{lin2021end-to-end} & 10.4 & - \\
Hasson \textit{et al.} \cite{hasson19_obman} & 11.0 & - \\
Hampali \textit{et al.} \cite{hampali2020honnotate} & 10.6 & 83.4 \\
Liu \textit{et al.} \cite{liu2021semi} & 9.8 & - \\
HandOccNet \cite{Park_2022_CVPR_HandOccNet} & \textbf{8.8}  & - \\
THOR-Net (Ours) & 10.7 & 26.3 \\
\hline
ArtiBoost \cite{li2021artiboost} & 10.4 & - \\
THOR-Net (Ours) & \textbf{10.0} & 23.7\\
\hline
\end{tabular}
\end{center}
\caption{\textbf{Comparison with state-of-the-art methods for 3D hand shape estimation on the HO-3D (v2) (upper table) and (v3) (lower table). Shown results are the Procrustes-aligned and non-aligned errors in \textit{mm}.} \vspace{-5mm}}
\label{tab:ho3d_shape}
\end{table}

\subsection{Ablation Study}

We study the impact of the proposed coarse-to-fine GraFormer on the hand shape reconstruction by evaluating three versions of the shape generator network. 
As shown in Figure \ref{fig:pipeline}, the coarse-to-fine network consists of three GraFormers. 
To show the effectiveness of this choice, we test the network with 1 GraFormer and 2 GraFormers.
Those two experiments are shown in Table \ref{tab:ablation} with IDs $1$ and $2$, respectively.
The results show that the network with three GraFormers achieves the best performance.
This shows that the gradual increase of the deep coarse-to-fine network is useful for shape estimation. 

%
%
%
To justify the choice of appending heatmaps to a feature vector of size $2048$ as the initial Coarse-to-fine graph, we test four other different graph input modalities. 
Instead of using the heatmaps, we try to use either the estimated 2D pose or the 3D pose.
%
%
In addition, we try two different feature vector sizes (i.e., $1024$ and $4096$) to test the capacity of the model and its correlation to the results. 
%
%
%
From the results shown in Table \ref{tab:ablation}, it is clear that the heatmap representation along with the feature vector of size $2048$ produces the best accuracy.

\begin{table}[t]
\begin{center}
\begin{tabular}{|c|c|c|c|c|}
\hline
ID & \# GraFo. & Gr. Inp. & $\mathcal{V}$ Al. Err. & $\mathcal{V}$ Err.  \\
\hline\hline
1 & 1 & $\mathcal{H}$ + $\mathcal{F}_{2048}$ & 11.4 & 26.8 \\
2 & 2 & $\mathcal{H}$ + $\mathcal{F}_{2048}$ & 11.1 & 26.4 \\
4 & 3 & 2D pose + $\mathcal{F}_{2048}$ & 14.6 & 46.8 \\
3 & 3 & 3D pose + $\mathcal{F}_{2048}$ & 10.9 & 27.0 \\ 
5 & 3 & $\mathcal{H}$ + $\mathcal{F}_{1024}$ & 13.5 & 29.5 \\
6 & 3 & $\mathcal{H}$ + $\mathcal{F}_{4096}$ & 11.8 & 28.4 \\
7 & 3 & $\mathcal{H}$ + $\mathcal{F}_{2048}$ & \textbf{10.0} & \textbf{23.7} \\
\hline
\end{tabular}
\end{center}
\caption{\textbf{Ablation study for the depth of the Coarse-to-fine shape generator and the graph input modality.}}
\label{tab:ablation}
\end{table}

\subsection{Quality of textured Shapes}

\begin{figure}[t]
\begin{center}
  \includegraphics[width=\linewidth]{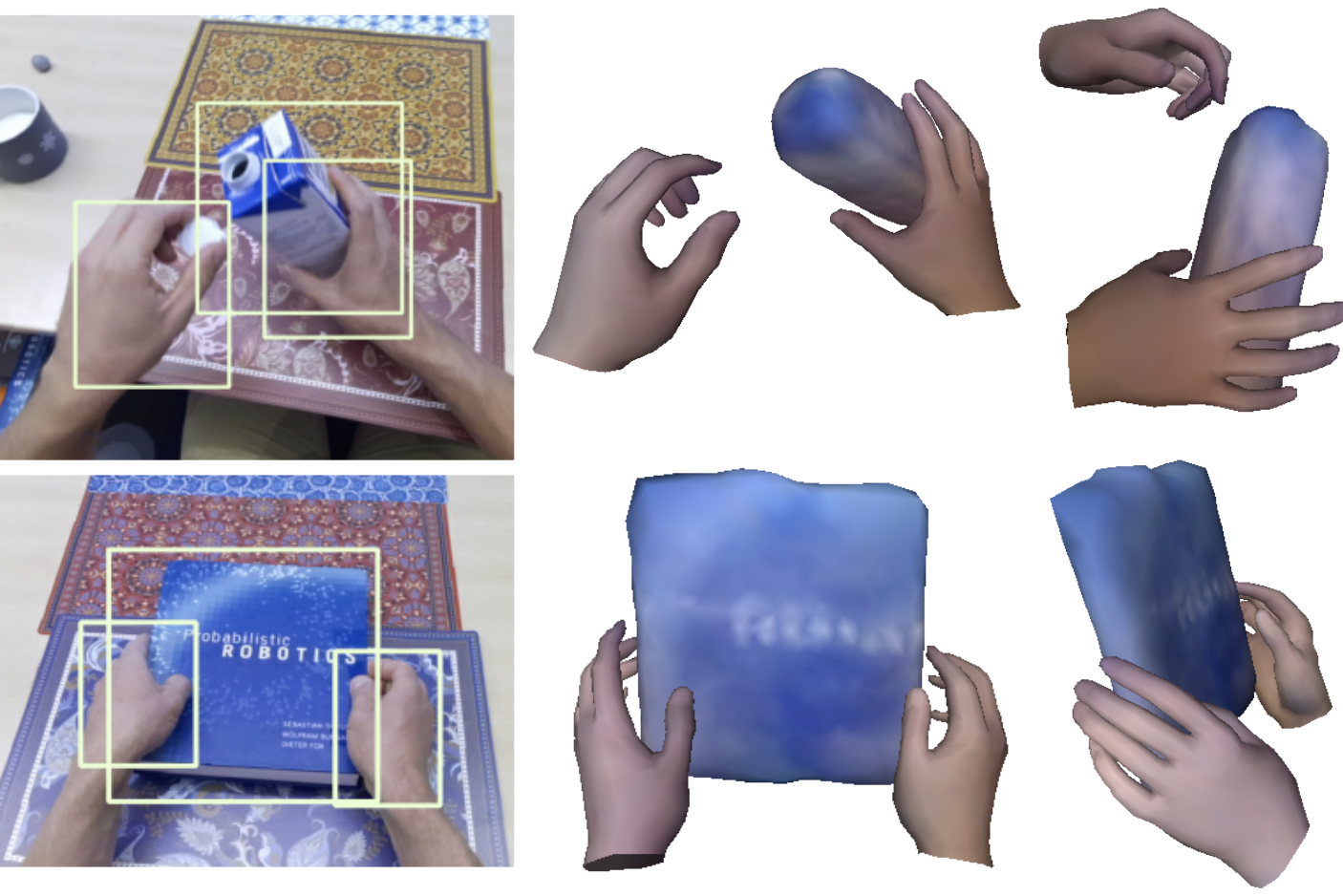}
\end{center}
\vspace{-3mm}
  \caption{Examples of textured hands and objects estimated using photometric loss. \vspace{-5mm}}

  \label{fig:texture}
\end{figure}

Figure \ref{fig:texture} shows the quality of the per-vertex texture values produced by the photometric loss. 
The network was able to reconstruct the hands despite the occlusions.
Furthermore, it managed to capture some of the text details and the blue color of the book.
Sometimes, a smoothing effect for the object colors happens as shown in the milk bottle hiding the details.
In addition, the hands are affected by lighting conditions which cause different tones of skin color.
The object's edges have artifacts as shown in the blue book.

\section{Conclusion and Future Work}

In this work, we propose THOR-Net to reconstruct realistic two hands interacting with an object from a monocular RGB frame.
The network consists of two stages: 2D feature extraction using Keypoint RCNN and 3D reconstruction using a coarse-to-fine GraFormer network. 
To obtain a per-vertex texture for shapes, we train the network using self-supervised photometric loss. 
The quantitative and qualitative evaluation shows the effectiveness of our coarse-to-fine network in two-hands-object shape estimation compared to previous methods. 
Qualitative results of photometric loss show a smoothing effect in texture estimation which suggests more future investigation.
For future work, temporal constraints from videos can be leveraged to model a Spatio-temporal graph that can improve the reconstructions.
\vspace{-5mm}
\paragraph{Acknowledgments}
This work has been partially funded by the German BMBF project GreifbAR (Grant Nr 16SV8732), and by the EU project FLUENTLY (Grant Nr 101058680).

\pagebreak

{\small
\bibliographystyle{ieee_fullname}
\bibliography{egbib}
}

\end{document}